\pdfoutput=1

\documentclass[11pt]{article}
\usepackage[table,xcdraw]{xcolor}

\usepackage{acl}

\usepackage{times}
\usepackage{latexsym}
\usepackage{multirow}
\usepackage{todonotes}
\usepackage{amssymb}

\usepackage[T1]{fontenc}

\usepackage[utf8]{inputenc}

\usepackage{microtype}

\usepackage{booktabs}

\title{Zero-Shot Dependency Parsing with Worst-Case Aware \\ Automated Curriculum Learning}

\author{Miryam de Lhoneux$^{1,2,3}$, 
Sheng Zhang$^4$,  Anders Søgaard$^1$  \\
          $^1$University of Copenhagen, Denmark
           $^2$Uppsala University, Sweden 
           $^3$KU Leuven, Belgium \\
           $^4$National University of Defense Technology, China\\
           \texttt{\{ml,soegaard\}@di.ku.dk, zhangsheng@nudt.edu.cn}}

\begin{document}
\maketitle
\begin{abstract}
  Large multilingual pretrained language models such as mBERT and XLM-RoBERTa have been found to be surprisingly effective for cross-lingual transfer of syntactic parsing models \citep{wu-dredze-2019-beto}, but only between related languages. However, source and training languages are rarely related, when parsing truly low-resource languages. To close this gap, we adopt a method from multi-task learning, which relies on automated curriculum learning, to dynamically optimize for parsing performance on {\em outlier} languages. We show that this approach is significantly better than uniform and size-proportional sampling in the zero-shot setting. 
\end{abstract}

\section{Introduction}
The field of multilingual NLP is booming \cite{10.1162/coli_r_00372}. This is due in no small part to large multilingual pretrained language models (PLMs) such as mBERT \citep{devlin-etal-2019-bert} and XLM-RoBERTa \citep{conneau-etal-2020-unsupervised}, which have been found to have surprising cross-lingual transfer capabilities in spite of receiving no cross-lingual supervision.\footnote{In the early days, cross-lingual transfer for dependency parsing relied on projection across word alignments \cite{spreyer-kuhn-2009-data,agic-etal-2016-multilingual} or {\em delexicalized transfer} of abstract syntactic features \cite{zeman-resnik-2008-cross,mcdonald-etal-2011-multi,sogaard-2011-data,cohen-etal-2011-unsupervised}. Delexicalized transfer was later 're-lexicalized' by word clusters \cite{tackstrom-etal-2012-cross} and word embeddings \cite{duong-etal-2015-cross}, but with the introduction of multilingual contextualized language models, transfer models no longer rely on abstract syntactic features, removing an important bottleneck for transfer approaches to scale to truly low-resource languages.} 
\citet{wu-dredze-2019-beto}, for example, found mBERT to perform well in a zero-shot setting when fine-tuned for five different NLP tasks in different languages. There is, however, a sharp divide between languages that benefit from this transfer and languages that do not, and there is ample evidence that transfer works best between typologically similar languages \citep[][among others]{pires-etal-2019-multilingual,lauscher-etal-2020-zero}. This means that the majority of world languages that are \emph{truly low-resource} are still left behind and inequalities in access to language technology are increasing. 

Large multilingual PLMs are typically fine-tuned using training data from a sample of languages that is supposed to be representative of the languages that the models are later applied to. However, this is difficult to achieve in practice, as multilingual datasets are not well balanced for typological diversity and contain a skewed distribution of typological features \citep{ponti-etal-2021-minimax}. This problem can be mitigated by using methods that sample from skewed distributions in a way that is robust to outliers.

\citet{zhang2020worst} recently developed such a method. It uses curriculum learning with a worst-case-aware loss for multi-task learning. They trained their model on a subset of the GLUE benchmark \citep{wang-etal-2018-glue} and tested on outlier tasks. This led to improved zero-shot performance on these outlier tasks. This method can be applied to multilingual NLP where different languages are considered different tasks. This is what we do in this work, for the case of multilingual dependency parsing. 
Multilingual dependency parsing is an ideal test case for this method, as the Universal Dependency treebanks \citep{nivre-etal-2020-universal} are currently the manually annotated dataset that covers the most typological diversity \citep{ponti-etal-2021-minimax}. 

Our research question can be formulated as such:
  {\em Can worst-case aware automated curriculum learning improve zero-shot cross-lingual dependency parsing?}\footnote{Our work is related to work in meta-learning for zero-shot cross-lingual transfer, in particular \citet{ponti-etal-2021-minimax}, who use worst-case-aware meta-learning to find good initializations for target languages. \citet{ponti-etal-2021-minimax} report zero-shot results for cross-lingual part-of-speech tagging and question-answering, with error reductions comparable to ours. Meta-learning also has been used for zero-shot cross-lingual learning by others \citep{nooralahzadeh-etal-2020-zero,xu-etal-2021-soft}, but using average loss rather than worst-case-aware objectives.}

\section{Worst-Case-Aware Curriculum Learning}
In multi-task learning, the total loss is generally the average of losses of different tasks:

\begin{equation}
    \min_\theta \ell(\theta) = \min_\theta \frac{1}{n}\sum_{i=1}^n{\ell_i(\theta)}
\end{equation}
where $l_i$ is the loss of task $i$. The architecture we use in this paper is adapted from \citet{zhang2020worst}, which is an automated curriculum learning \cite{graves2017automated} framework to learn a worst-case-aware loss in a multi-task learning scenario. The architecture consists of a sampler, a buffer, a trainer and a multilingual dependency parsing model. The two main components are the sampler, which adopts a curriculum sampling strategy to dynamically sample data batches, and the trainer which uses worst-case-aware strategy to train the model. The framework repeats the following steps: (1) the sampler samples data batches of different languages to the buffer; (2) the trainer uses a worst-case strategy to train the model; (3) the automated curriculum learning strategy of the sampler is updated.

\paragraph{Sampling data batches} We view multilingual dependency parsing as multi-task learning where parsing in each individual language is considered a task. This means that the target of the sampler at each step is to choose a data batch from one language.
This is a typical multi-arm bandit problem \citep{even2002pac}. The sampler should choose bandits that have higher rewards, and in our scenario, 
data batches that have a higher loss on the model are more likely to be selected by the sampler and therefore, in a later stage, used by the trainer.
Automated curriculum learning is adopted to push a batch with its loss into the buffer at each time step.
The buffer consists of $n$ first-in-first-out queues, and each queue corresponds to a task (in our case, a language). The procedure repeats $k$ times and, at each round, $k$ data batches 
are pushed into the buffer.

\paragraph{Worst-case-aware risk minimization}

In multilingual and multi-task learning scenarios, in which we jointly minimize our risk across $n$ languages or tasks, we are confronted with the question of how to summarize $n$ losses. In other words, the question is how to compare two loss vectors $\alpha$ and $\beta$ containing losses for all tasks $l_i,\ldots l_n$: 

$$\alpha=[\ell^1_1,\ldots,\ell^1_n]$$
and 
$$\beta=[\ell^2_1,\ldots,\ell^2_n]$$

The most obvious thing to do is to minimize the mean of the $n$ losses, asking whether ${\sum_{\ell\in\alpha}\ell}<{\sum_{\ell\in\beta}\ell}$. We could also, motivated by robustness \cite{sogaard-2013-part} and fairness \cite{pmlr-v97-williamson19a}, minimize the maximum (supremum) of the $n$ losses, asking whether ${\max_{\ell\in\alpha}\ell}<{\max_{\ell\in\beta}\ell}$. \citet{mehta2012minimax} observed that these two loss summarizations are extremes that can be generalized by a family of multi-task loss functions that summarize the loss of $n$ tasks as the $L^p$ norm of the $n$-dimensional loss vector. Minimizing the average loss then corresponds to computing the $L^1$ norm, i.e., asking whether $|\alpha|^1<|\beta|^1$, and minimizing the worst-case loss corresponds to computing the $L^\infty$ (supremum) norm, i.e., asking whether $|\alpha|^\infty<|\beta|^\infty$. 

\citet{zhang2020worst} present a stochastic generalization of the $L^\infty$ loss summarization and a practical approach to minimizing this family of losses through automated curriculum learning \cite{graves2017automated}: The core idea behind their generalization is to optimize the worst-case loss with a certain probability, otherwise optimize the average (loss-proportional) loss with the remaining probability. The hyperparameter $\phi$ is introduced by the worst-case-aware risk minimization to trade off the balance between the worst-case and the loss-proportional losses. The loss family is formally defined as:

\begin{small}
\begin{equation}
    \min \ell(\theta)
    =\left\{
\begin{array}{rl}
\min \: \max_i (\ell_i(\theta)), &  p < \phi \\
\\
\min \ell_{\tilde{i}}(\theta), &  p \geq \phi, \tilde{i} \sim P_\ell \\
\end{array} \right.
\end{equation}
\end{small} 
where $p \in [0,1]$ is a random generated rational number, and  $P_\ell=\frac{\ell_i}{\sum_{j\leq n}{\ell_j}}$ is the normalized probability distribution of task losses. If $ p < \phi$ the model chooses the maximum loss among all tasks, otherwise, it randomly chooses one loss according to the loss distribution. If the hyperparameter $\phi$ equals $1$, the trainer updates the model with respect to the worst-case loss. On the contrary, if $\phi=0$, the trainer loss-proportionally samples one loss.

\paragraph{Sampling strategy updates} The model updates its parameters with respect to the loss chosen by the trainer. After that, the sampler updates its policy according to the behavior of the trainer. At each round, the policy of the task that is selected by the trainer receives positive rewards and the policy of all other tasks that have been selected by the sampler receive negative rewards.

\paragraph{The multilingual dependency parsing model} 

We use a standard biaffine graph-based dependency parser \citep{dozat16biaffine}. The model takes token representations of words from a contextualized language model (mBERT or XLM-R) as input and classifies head and dependency relations between words in the sentence. 
The Chu-Liu-Edmonds algorithm \citep{chu65,edmonds67} is then used to decode the score matrix into a tree. 
All languages share the same encoder and decoder in order to learn features from different languages, and more importantly to perform zero-shot transfer to unseen languages. 

\section{Experiments}

We base our experimental design on \citet{ustun-etal-2020-udapter}, a recent paper doing zero-shot dependency parsing with good performance on a large number of languages. They fine-tune mBERT for dependency parsing using training data from a sample of 13 typologically diverse languages from Universal Dependencies \citep[UD;][]{nivre-etal-2020-universal}, listed in Table~\ref{tab:13lang}. 
For testing, they use 30 test sets from treebanks whose language has not been seen at fine-tuning time. We use the same training and test sets and experiment both with mBERT and XLM-R as PLMs. It is important to note that not all of the test languages have been seen by the PLMs.\footnote{Information about which treebank has been seen by which PLM can be found in Appendix~\ref{sec:res_by_tb}.}

\begin{table}[t]
  \small
\begin{tabular}{llll}
Language & Treebank  & Genus    & Lang. family \\
\toprule
Arabic   & PADT      & Semitic  & Afro-Asiatic \\
Basque   & BDT       & Basque   & Basque       \\
Chinese  & GSD       & Chinese  & Sino-Tibetan \\
English  & EWT       & Germanic & IE           \\
Finnish  & TDT       & Finnic   & Uralic       \\
Hebrew   & HTB       & Semitic  & Afro-Asiatic \\
Hindi    & HDTB      & Indic    & IE           \\
Italian  & ISDT      & Romance  & IE           \\
Japanese & GSD       & Japanese & Japanese     \\
Korean   & GSD       & Korean   & Korean       \\
Russian  & SynTagRus & Slavic   & IE           \\
Swedish  & Talbanken & Germanic & IE           \\
Turkish  & IMST      & Turkic   & Altaic       
\end{tabular}
\caption{13 training treebanks. IE=Indo-European.}
\label{tab:13lang}
\end{table}

We test worst-case aware learning with different values of $\phi$ and compare this to three main baselines:
\textit{size-proportional} samples batches proportionally to the data sizes of the training treebanks,
\textit{uniform} samples from different treebanks with equal probability, thereby effectively reducing the size of the training data,
and \textit{smooth-sampling}
uses the smooth sampling method developed in \citet{van-der-goot-etal-2021-massive} which samples from multiple languages using a multinomial distribution. 
These baselines are competitive with the state-of-the-art when using mBERT, they are within 0.2 to 0.4 LAS points from the baseline of \citet{ustun-etal-2020-udapter} on the same test sets. When using XLM-R, they are largely above the state-of-the-art.

We implement all models using MaChAmp \citep{van-der-goot-etal-2021-massive}, a library for multi-task learning based on AllenNLP \citep{gardner-etal-2018-allennlp}. The library uses transformers from HuggingFace \citep{wolf-etal-2020-transformers}.
Our code is publicly available.\footnote{\url{https://github.com/mdelhoneux/machamp-worst_case_acl}}

Our main results are in Table~\ref{tab:main_res} where we report average scores across test sets, for space reasons. Results broken down by test treebank can be found in Table~\ref{tab:res_all} in Appendix~\ref{sec:res_by_tb}.
We can see that worst-case-aware training outperforms all of our baselines in the zero-shot setting, highlighting the effectiveness of this method. This answers positively our research question \emph{Can worst-case aware automated curriculum learning improve zero-shot dependency parsing?}

Our results using mBERT are more than 1 LAS point above the corresponding baselines. Our best model is significantly better than the best baseline with $p<.01$ according to a bootstrap test across test treebanks.
Our best model with mBERT comes close to Udapter (36.5 LAS on the same test sets) while being a lot simpler and not using external resources such as typological features, which are not always available for truly low-resource languages.

The results with XLM-R are much higher in general\footnote{Note, however, that the results are not directly comparable since different subsets of test languages have been seen by the two PLMs.}
  but the trends are similar: all our models outperform all of our baselines albeit with smaller differences. There is only a 0.4 LAS difference between our best model and the best baseline, but it is still significant with $p<.05$ according to a bootstrap test across test treebanks.
  This highlights the robustness of the XLM-R model itself. Our results with XLM-R outperform Udapter by close to 7 LAS points. 

\begin{table}[t]
\centering
\begin{tabular}{clrr}
    &              & \multicolumn{1}{l}{mBERT} & \multicolumn{1}{l}{XLM-R} \\
                  \toprule
\multirow{3}{*}{\rotatebox[origin=c]{90}{\scriptsize \sc Ours}}&$\phi$=0             
& \textbf{36.4}         &42.1        \\
&$\phi$=0.5           
& 36.1                  &{\bf 42.3}        \\
&$\phi$=1            
& 36.1              &{\bf 42.3}            \\
\midrule
\multirow{3}{*}{\rotatebox[origin=c]{90}{\scriptsize \sc Baselines}}&size-proportional 
& 35.0                  &41.9        \\
&smooth-sampling 
& 35.2              &41.7            \\
&uniform           
& 35.2      &41.4\\
\bottomrule                   
\end{tabular}
\caption{{\bf Zero-shot performance:} Average LAS scores on the test sets of the 30 unseen (zero-shot) languages in the language split from \citet{ustun-etal-2020-udapter}.}
\label{tab:main_res}
\end{table}

\section{Varying the homogeneity of training samples}
We investigate the interaction between the effectiveness of worst-case learning and the representativeness of the sample of training languages.
It is notoriously difficult to construct a sample of treebanks that is representative of the languages in UD \citep{delhoneux17old,schluter-agic-2017-empirically,delhoneux19}. 
We can, however, easily construct samples that are \textbf{not} representative, for example, by taking a sample of related languages. We expect worst-case aware learning to lead to larger improvements in cases where some language types are underrepresented in the sample. We can construct an extreme case of underrepresentation by selecting a sample of training languages that has one or more clear outliers. For example we can construct a sample of related languages, add a single unrelated language in the mix, and then evaluate on other unrelated languages. We also expect that with a typologically diverse set of training languages, worst-case aware learning should lead to larger relative improvements than with a homogeneous sample, but perhaps slightly smaller improvements than with a very skewed sample.

We test these hypotheses by constructing seven samples of training languages in addition to the one used so far (\textsc{13lang}). We construct three different homogeneous samples using treebanks from three different genera: \textsc{germanic}, \textsc{romance} and \textsc{slavic}. We construct four skewed samples using the sample of romance languages and a language from a different language family, an \emph{outlier} language: Basque (eu), Arabic (ar), Turkish (tr) and Chinese (zh). Since we keep the sample of test sets constant, we do not include training data from languages that are in the test sets. 
The details of which treebanks are used for each of these samples can be found in Table~\ref{tab:training_samples} in Appendix~\ref{sec:training_samples}.

\begin{table}[t]
\begin{tabular}{l|rrrr}
sample            & \multicolumn{1}{l}{\textsc{base}} & \multicolumn{1}{l}{\textsc{ours}} & \multicolumn{1}{l}{$\delta$} & \multicolumn{1}{l}{RER}     \\ \hline
\textsc{13lang}   & 35.2                                  & 36.4                                  & \cellcolor[HTML]{AEDEC6}1.2  & \cellcolor[HTML]{78C9A1}1.9 \\ \hline
\textsc{germanic} & 30.7                                  & 31.4                                  & \cellcolor[HTML]{D7EFE3}0.7  & \cellcolor[HTML]{BDE5D1}1.0 \\
\textsc{slavic}   & 30.4                                  & 31.7                                  & \cellcolor[HTML]{A5DBC1}1.3  & \cellcolor[HTML]{77C8A0}1.9 \\
\textsc{romance}  & 31.3                                  & 32.5                                  & \cellcolor[HTML]{AEDEC6}1.2  & \cellcolor[HTML]{81CCA7}1.7 \\ \hline
\textsc{rom+eu}   & 33.3                                  & 34.8                                  & \cellcolor[HTML]{95D4B5}1.5  & \cellcolor[HTML]{57BB8A}2.2 \\
\textsc{rom+ar}   & 32.0                                  & 32.2                                  & \cellcolor[HTML]{FFFFFF}0.2  & \cellcolor[HTML]{F8FCFA}0.3 \\
\textsc{rom+tr}   & 32.2                                  & 33.0                                  & \cellcolor[HTML]{CEECDD}0.8  & \cellcolor[HTML]{AFDFC8}1.2 \\
\textsc{rom+zh}   & 33.4                                  & 34.1                                  & \cellcolor[HTML]{D7EFE3}0.7  & \cellcolor[HTML]{BAE3CF}1.1
\end{tabular}
\caption{LAS of best baseline (\textsc{base}) and best worst-case training (\textsc{ours}) when using mBERT as a PLM. Absolute difference ($\delta$) and relative error reduction (RER) between \textsc{ours} and \textsc{base}.}
\label{tab:diff_samples}
\end{table}

Results are in Table~\ref{tab:diff_samples} where we report the average LAS scores of our best model (out of the ones trained with the three different $\phi$ values) to the best of the three baselines. 
We can see first that, as expected, our typologically diverse sample performs best overall. This indicates that it is a good sample. 
We can also see that, as expected, the method works best with a skewed sample: the largest gains from using worst-case learning, both in terms of absolute LAS difference and relative error reduction, are seen for a skewed sample (\textsc{rom+eu}). However, contrary to expectations, the lowest gains are obtained for another skewed sample (\textsc{rom+ar}). The gains are also low for \textsc{rom+tr}, \textsc{rom+zh} and for \textsc{germanic}.
Additionally, there are slightly more gains from using worst-case aware learning with the \textsc{slavic} sample than for our typologically diverse sample. 
These results could be due to the different scripts of the languages involved both in training and testing. 

Looking at results of the different models on individual test languages (see Figure~\ref{fig:tb_res_plot} in Appendix~\ref{sec:tb_res_sample}), we find no clear pattern of the settings in which this method works best. We do note that the method always hurts Belarusian, which is perhaps unsurprising given that it is the test treebank for which the baseline is highest. Worst-case aware learning hurts Belarusian the least when using the \textsc{slavic} sample, indicating that, when using the other samples, the languages related to Belarusian are likely downsampled in favour of languages unrelated to it. 
Worst-case learning consistently helps Breton and Swiss German, indicating that the method might work best for languages that are underrepresented within their language family but not necessarily outside of it. 
For Swiss German, worst-case learning helps least when using the \textsc{germanic} sample where it is less of an outlier.

\section{Conclusion}
In this work, we have adopted a method from multi-task learning which relies on automated curriculum learning to the case of multilingual dependency parsing. This method allows to dynamically optimize for parsing performance on \emph{outlier} languages. We found this method to improve dependency parsing on a sample of 30 test languages in the zero-shot setting, compared to sampling data uniformly across treebanks from different languages, or proportionally to the size of the treebanks.
We investigated the impact of varying the homogeneity of the sample of training treebanks on the usefulness of the method and found conflicting evidence with different samples.
This leaves open questions about the relationship between the languages used for training and the ones used for testing. 

\section{Acknowledgements}
We thank Daniel Hershcovich and Ruixiang Cui for comments on a draft of the paper, as well as the members of CoAStaL for discussions about the content of the paper.
Miryam de Lhoneux was funded by the Swedish Research Council (grant 2020-00437). Anders S{\o}gaard was funded by the Innovation Fund Denmark and a Google Focused Research Award. 
We acknowledge the computational resources provided by CSC in Finland through NeIC-NLPL and the EOSC-Nordic NLPL use case (www.nlpl.eu).

\bibliography{main,custom}

\begin{thebibliography}{35}
\expandafter\ifx\csname natexlab\endcsname\relax\def\natexlab#1{#1}\fi

\bibitem[{Agi{\'c} et~al.(2016)Agi{\'c}, Johannsen, Plank,
  Mart{\'\i}nez~Alonso, Schluter, and S{\o}gaard}]{agic-etal-2016-multilingual}
{\v{Z}}eljko Agi{\'c}, Anders Johannsen, Barbara Plank, H{\'e}ctor
  Mart{\'\i}nez~Alonso, Natalie Schluter, and Anders S{\o}gaard. 2016.
\newblock \href {https://doi.org/10.1162/tacl_a_00100} {Multilingual projection
  for parsing truly low-resource languages}.
\newblock \emph{Transactions of the Association for Computational Linguistics},
  4:301--312.

\bibitem[{Agirre(2020)}]{10.1162/coli_r_00372}
Eneko Agirre. 2020.
\newblock \href {https://doi.org/10.1162/coli_r_00372} {{Cross-Lingual Word
  Embeddings}}.
\newblock \emph{Computational Linguistics}, 46(1):245--248.

\bibitem[{Chu and Liu(1965)}]{chu65}
Yoeng-Jin Chu and Tseng-hong Liu. 1965.
\newblock On the shortest arborescence of a directed graph.
\newblock \emph{Scientia Sinica}, 14:1396--1400.

\bibitem[{Cohen et~al.(2011)Cohen, Das, and
  Smith}]{cohen-etal-2011-unsupervised}
Shay~B. Cohen, Dipanjan Das, and Noah~A. Smith. 2011.
\newblock \href {https://aclanthology.org/D11-1005} {Unsupervised structure
  prediction with non-parallel multilingual guidance}.
\newblock In \emph{Proceedings of the 2011 Conference on Empirical Methods in
  Natural Language Processing}, pages 50--61, Edinburgh, Scotland, UK.
  Association for Computational Linguistics.

\bibitem[{Conneau et~al.(2020)Conneau, Khandelwal, Goyal, Chaudhary, Wenzek,
  Guzm{\'a}n, Grave, Ott, Zettlemoyer, and
  Stoyanov}]{conneau-etal-2020-unsupervised}
Alexis Conneau, Kartikay Khandelwal, Naman Goyal, Vishrav Chaudhary, Guillaume
  Wenzek, Francisco Guzm{\'a}n, Edouard Grave, Myle Ott, Luke Zettlemoyer, and
  Veselin Stoyanov. 2020.
\newblock \href {https://doi.org/10.18653/v1/2020.acl-main.747} {Unsupervised
  cross-lingual representation learning at scale}.
\newblock In \emph{Proceedings of the 58th Annual Meeting of the Association
  for Computational Linguistics}, pages 8440--8451, Online. Association for
  Computational Linguistics.

\bibitem[{de~Lhoneux(2019)}]{delhoneux19}
Miryam de~Lhoneux. 2019.
\newblock \href
  {http://www.diva-portal.org/smash/get/diva2:1357373/FULLTEXT01.pdf}
  {\emph{{Linguistically Informed Neural Dependency Parsing for Typologically
  Diverse Languages}}}.
\newblock Ph.D. thesis, Uppsala University.

\bibitem[{de~Lhoneux et~al.(2017)de~Lhoneux, Stymne, and
  Nivre}]{delhoneux17old}
Miryam de~Lhoneux, Sara Stymne, and Joakim Nivre. 2017.
\newblock {Old School vs. New School: Comparing Transition-Based Parsers with
  and without Neural Network Enhancement.}
\newblock In \emph{Proceedings of the 15th Treebanks and Linguistic Theories
  Workshop (TLT)}, pages 99--110.

\bibitem[{Devlin et~al.(2019)Devlin, Chang, Lee, and
  Toutanova}]{devlin-etal-2019-bert}
Jacob Devlin, Ming-Wei Chang, Kenton Lee, and Kristina Toutanova. 2019.
\newblock \href {https://doi.org/10.18653/v1/N19-1423} {{BERT}: Pre-training of
  deep bidirectional transformers for language understanding}.
\newblock In \emph{Proceedings of the 2019 Conference of the North {A}merican
  Chapter of the Association for Computational Linguistics: Human Language
  Technologies, Volume 1 (Long and Short Papers)}, pages 4171--4186,
  Minneapolis, Minnesota. Association for Computational Linguistics.

\bibitem[{Dozat and Manning(2017)}]{dozat16biaffine}
Timothy Dozat and Christopher Manning. 2017.
\newblock {Deep Biaffine Attention for Neural Dependency Parsing}.
\newblock In \emph{{Proceedings of the 5th International Conference on Learning
  Representations.}}

\bibitem[{Duong et~al.(2015)Duong, Cohn, Bird, and
  Cook}]{duong-etal-2015-cross}
Long Duong, Trevor Cohn, Steven Bird, and Paul Cook. 2015.
\newblock \href {https://doi.org/10.18653/v1/K15-1012} {Cross-lingual transfer
  for unsupervised dependency parsing without parallel data}.
\newblock In \emph{Proceedings of the Nineteenth Conference on Computational
  Natural Language Learning}, pages 113--122, Beijing, China. Association for
  Computational Linguistics.

\bibitem[{Edmonds(1967)}]{edmonds67}
Jack Edmonds. 1967.
\newblock {Optimum Branchings}.
\newblock \emph{Journal of Research of the National Bureau of Standards},
  71B:233--240.

\bibitem[{Even-Dar et~al.(2002)Even-Dar, Mannor, and Mansour}]{even2002pac}
Eyal Even-Dar, Shie Mannor, and Yishay Mansour. 2002.
\newblock Pac bounds for multi-armed bandit and markov decision processes.
\newblock In \emph{International Conference on Computational Learning Theory},
  pages 255--270. Springer.

\bibitem[{Gardner et~al.(2018)Gardner, Grus, Neumann, Tafjord, Dasigi, Liu,
  Peters, Schmitz, and Zettlemoyer}]{gardner-etal-2018-allennlp}
Matt Gardner, Joel Grus, Mark Neumann, Oyvind Tafjord, Pradeep Dasigi,
  Nelson~F. Liu, Matthew Peters, Michael Schmitz, and Luke Zettlemoyer. 2018.
\newblock \href {https://doi.org/10.18653/v1/W18-2501} {{A}llen{NLP}: A deep
  semantic natural language processing platform}.
\newblock In \emph{Proceedings of Workshop for {NLP} Open Source Software
  ({NLP}-{OSS})}, pages 1--6, Melbourne, Australia. Association for
  Computational Linguistics.

\bibitem[{Graves et~al.(2017)Graves, Bellemare, Menick, Munos, and
  Kavukcuoglu}]{graves2017automated}
Alex Graves, Marc~G. Bellemare, Jacob Menick, Remi Munos, and Koray
  Kavukcuoglu. 2017.
\newblock \href {http://arxiv.org/abs/1704.03003} {Automated curriculum
  learning for neural networks}.

\bibitem[{Lauscher et~al.(2020)Lauscher, Ravishankar, Vuli{\'c}, and
  Glava{\v{s}}}]{lauscher-etal-2020-zero}
Anne Lauscher, Vinit Ravishankar, Ivan Vuli{\'c}, and Goran Glava{\v{s}}. 2020.
\newblock \href {https://doi.org/10.18653/v1/2020.emnlp-main.363} {From zero to
  hero: {O}n the limitations of zero-shot language transfer with multilingual
  {T}ransformers}.
\newblock In \emph{Proceedings of the 2020 Conference on Empirical Methods in
  Natural Language Processing (EMNLP)}, pages 4483--4499, Online. Association
  for Computational Linguistics.

\bibitem[{McDonald et~al.(2011)McDonald, Petrov, and
  Hall}]{mcdonald-etal-2011-multi}
Ryan McDonald, Slav Petrov, and Keith Hall. 2011.
\newblock \href {https://aclanthology.org/D11-1006} {Multi-source transfer of
  delexicalized dependency parsers}.
\newblock In \emph{Proceedings of the 2011 Conference on Empirical Methods in
  Natural Language Processing}, pages 62--72, Edinburgh, Scotland, UK.
  Association for Computational Linguistics.

\bibitem[{Mehta et~al.(2012)Mehta, Lee, and Gray}]{mehta2012minimax}
Nishant Mehta, Dongryeol Lee, and Alexander~G Gray. 2012.
\newblock Minimax multi-task learning and a generalized loss-compositional
  paradigm for mtl.
\newblock In \emph{Advances in Neural Information Processing Systems}, pages
  2150--2158.

\bibitem[{Nivre et~al.(2020)Nivre, de~Marneffe, Ginter, Haji{\v{c}}, Manning,
  Pyysalo, Schuster, Tyers, and Zeman}]{nivre-etal-2020-universal}
Joakim Nivre, Marie-Catherine de~Marneffe, Filip Ginter, Jan Haji{\v{c}},
  Christopher~D. Manning, Sampo Pyysalo, Sebastian Schuster, Francis Tyers, and
  Daniel Zeman. 2020.
\newblock \href {https://aclanthology.org/2020.lrec-1.497} {{U}niversal
  {D}ependencies v2: An evergrowing multilingual treebank collection}.
\newblock In \emph{Proceedings of the 12th Language Resources and Evaluation
  Conference}, pages 4034--4043, Marseille, France. European Language Resources
  Association.

\bibitem[{Nooralahzadeh et~al.(2020)Nooralahzadeh, Bekoulis, Bjerva, and
  Augenstein}]{nooralahzadeh-etal-2020-zero}
Farhad Nooralahzadeh, Giannis Bekoulis, Johannes Bjerva, and Isabelle
  Augenstein. 2020.
\newblock \href {https://doi.org/10.18653/v1/2020.emnlp-main.368} {Zero-shot
  cross-lingual transfer with meta learning}.
\newblock In \emph{Proceedings of the 2020 Conference on Empirical Methods in
  Natural Language Processing (EMNLP)}, pages 4547--4562, Online. Association
  for Computational Linguistics.

\bibitem[{Pires et~al.(2019)Pires, Schlinger, and
  Garrette}]{pires-etal-2019-multilingual}
Telmo Pires, Eva Schlinger, and Dan Garrette. 2019.
\newblock \href {https://doi.org/10.18653/v1/P19-1493} {How multilingual is
  multilingual {BERT}?}
\newblock In \emph{Proceedings of the 57th Annual Meeting of the Association
  for Computational Linguistics}, pages 4996--5001, Florence, Italy.
  Association for Computational Linguistics.

\bibitem[{Ponti et~al.(2021)Ponti, Aralikatte, Shrivastava, Reddy, and
  S{\o}gaard}]{ponti-etal-2021-minimax}
Edoardo~Maria Ponti, Rahul Aralikatte, Disha Shrivastava, Siva Reddy, and
  Anders S{\o}gaard. 2021.
\newblock \href {https://doi.org/10.18653/v1/2021.findings-acl.106} {Minimax
  and neyman{--}{P}earson meta-learning for outlier languages}.
\newblock In \emph{Findings of the Association for Computational Linguistics:
  ACL-IJCNLP 2021}, pages 1245--1260, Online. Association for Computational
  Linguistics.

\bibitem[{Schluter and Agi{\'c}(2017)}]{schluter-agic-2017-empirically}
Natalie Schluter and {\v{Z}}eljko Agi{\'c}. 2017.
\newblock \href {https://aclanthology.org/W17-0415} {Empirically sampling
  {U}niversal {D}ependencies}.
\newblock In \emph{Proceedings of the {N}o{D}a{L}i{D}a 2017 Workshop on
  Universal Dependencies ({UDW} 2017)}, pages 117--122, Gothenburg, Sweden.
  Association for Computational Linguistics.

\bibitem[{S{\o}gaard(2011)}]{sogaard-2011-data}
Anders S{\o}gaard. 2011.
\newblock \href {https://aclanthology.org/P11-2120} {Data point selection for
  cross-language adaptation of dependency parsers}.
\newblock In \emph{Proceedings of the 49th Annual Meeting of the Association
  for Computational Linguistics: Human Language Technologies}, pages 682--686,
  Portland, Oregon, USA. Association for Computational Linguistics.

\bibitem[{S{\o}gaard(2013)}]{sogaard-2013-part}
Anders S{\o}gaard. 2013.
\newblock \href {https://aclanthology.org/P13-2113} {Part-of-speech tagging
  with antagonistic adversaries}.
\newblock In \emph{Proceedings of the 51st Annual Meeting of the Association
  for Computational Linguistics (Volume 2: Short Papers)}, pages 640--644,
  Sofia, Bulgaria. Association for Computational Linguistics.

\bibitem[{Spreyer and Kuhn(2009)}]{spreyer-kuhn-2009-data}
Kathrin Spreyer and Jonas Kuhn. 2009.
\newblock \href {https://aclanthology.org/W09-1104} {Data-driven dependency
  parsing of new languages using incomplete and noisy training data}.
\newblock In \emph{Proceedings of the Thirteenth Conference on Computational
  Natural Language Learning ({C}o{NLL}-2009)}, pages 12--20, Boulder, Colorado.
  Association for Computational Linguistics.

\bibitem[{T{\"a}ckstr{\"o}m et~al.(2012)T{\"a}ckstr{\"o}m, McDonald, and
  Uszkoreit}]{tackstrom-etal-2012-cross}
Oscar T{\"a}ckstr{\"o}m, Ryan McDonald, and Jakob Uszkoreit. 2012.
\newblock \href {https://aclanthology.org/N12-1052} {Cross-lingual word
  clusters for direct transfer of linguistic structure}.
\newblock In \emph{Proceedings of the 2012 Conference of the North {A}merican
  Chapter of the Association for Computational Linguistics: Human Language
  Technologies}, pages 477--487, Montr{\'e}al, Canada. Association for
  Computational Linguistics.

\bibitem[{{\"U}st{\"u}n et~al.(2020){\"U}st{\"u}n, Bisazza, Bouma, and van
  Noord}]{ustun-etal-2020-udapter}
Ahmet {\"U}st{\"u}n, Arianna Bisazza, Gosse Bouma, and Gertjan van Noord. 2020.
\newblock \href {https://doi.org/10.18653/v1/2020.emnlp-main.180} {{UD}apter:
  Language adaptation for truly {U}niversal {D}ependency parsing}.
\newblock In \emph{Proceedings of the 2020 Conference on Empirical Methods in
  Natural Language Processing (EMNLP)}, pages 2302--2315, Online. Association
  for Computational Linguistics.

\bibitem[{van~der Goot et~al.(2021)van~der Goot, {\"U}st{\"u}n, Ramponi,
  Sharaf, and Plank}]{van-der-goot-etal-2021-massive}
Rob van~der Goot, Ahmet {\"U}st{\"u}n, Alan Ramponi, Ibrahim Sharaf, and
  Barbara Plank. 2021.
\newblock \href {https://doi.org/10.18653/v1/2021.eacl-demos.22} {Massive
  choice, ample tasks ({M}a{C}h{A}mp): A toolkit for multi-task learning in
  {NLP}}.
\newblock In \emph{Proceedings of the 16th Conference of the European Chapter
  of the Association for Computational Linguistics: System Demonstrations},
  pages 176--197, Online. Association for Computational Linguistics.

\bibitem[{Wang et~al.(2018)Wang, Singh, Michael, Hill, Levy, and
  Bowman}]{wang-etal-2018-glue}
Alex Wang, Amanpreet Singh, Julian Michael, Felix Hill, Omer Levy, and Samuel
  Bowman. 2018.
\newblock \href {https://doi.org/10.18653/v1/W18-5446} {{GLUE}: A multi-task
  benchmark and analysis platform for natural language understanding}.
\newblock In \emph{Proceedings of the 2018 {EMNLP} Workshop {B}lackbox{NLP}:
  Analyzing and Interpreting Neural Networks for {NLP}}, pages 353--355,
  Brussels, Belgium. Association for Computational Linguistics.

\bibitem[{Williamson and Menon(2019)}]{pmlr-v97-williamson19a}
Robert Williamson and Aditya Menon. 2019.
\newblock \href {https://proceedings.mlr.press/v97/williamson19a.html}
  {Fairness risk measures}.
\newblock In \emph{Proceedings of the 36th International Conference on Machine
  Learning}, volume~97 of \emph{Proceedings of Machine Learning Research},
  pages 6786--6797. PMLR.

\bibitem[{Wolf et~al.(2020)Wolf, Debut, Sanh, Chaumond, Delangue, Moi, Cistac,
  Rault, Louf, Funtowicz, Davison, Shleifer, von Platen, Ma, Jernite, Plu, Xu,
  Le~Scao, Gugger, Drame, Lhoest, and Rush}]{wolf-etal-2020-transformers}
Thomas Wolf, Lysandre Debut, Victor Sanh, Julien Chaumond, Clement Delangue,
  Anthony Moi, Pierric Cistac, Tim Rault, Remi Louf, Morgan Funtowicz, Joe
  Davison, Sam Shleifer, Patrick von Platen, Clara Ma, Yacine Jernite, Julien
  Plu, Canwen Xu, Teven Le~Scao, Sylvain Gugger, Mariama Drame, Quentin Lhoest,
  and Alexander Rush. 2020.
\newblock \href {https://doi.org/10.18653/v1/2020.emnlp-demos.6} {Transformers:
  State-of-the-art natural language processing}.
\newblock In \emph{Proceedings of the 2020 Conference on Empirical Methods in
  Natural Language Processing: System Demonstrations}, pages 38--45, Online.
  Association for Computational Linguistics.

\bibitem[{Wu and Dredze(2019)}]{wu-dredze-2019-beto}
Shijie Wu and Mark Dredze. 2019.
\newblock \href {https://doi.org/10.18653/v1/D19-1077} {Beto, bentz, becas: The
  surprising cross-lingual effectiveness of {BERT}}.
\newblock In \emph{Proceedings of the 2019 Conference on Empirical Methods in
  Natural Language Processing and the 9th International Joint Conference on
  Natural Language Processing (EMNLP-IJCNLP)}, pages 833--844, Hong Kong,
  China. Association for Computational Linguistics.

\bibitem[{Xu et~al.(2021)Xu, Haider, Krone, and Mansour}]{xu-etal-2021-soft}
Weijia Xu, Batool Haider, Jason Krone, and Saab Mansour. 2021.
\newblock \href {https://doi.org/10.18653/v1/2021.metanlp-1.2} {Soft layer
  selection with meta-learning for zero-shot cross-lingual transfer}.
\newblock In \emph{Proceedings of the 1st Workshop on Meta Learning and Its
  Applications to Natural Language Processing}, pages 11--18, Online.
  Association for Computational Linguistics.

\bibitem[{Zeman and Resnik(2008)}]{zeman-resnik-2008-cross}
Daniel Zeman and Philip Resnik. 2008.
\newblock \href {https://aclanthology.org/I08-3008} {Cross-language parser
  adaptation between related languages}.
\newblock In \emph{Proceedings of the {IJCNLP}-08 Workshop on {NLP} for Less
  Privileged Languages}.

\bibitem[{Zhang et~al.(2020)Zhang, Zhang, Zhang, and
  S{\o}gaard}]{zhang2020worst}
Sheng Zhang, Xin Zhang, Weiming Zhang, and Anders S{\o}gaard. 2020.
\newblock Worst-case-aware curriculum learning for zero and few shot transfer.
\newblock \emph{arXiv preprint arXiv:2009.11138}.

\end{thebibliography}
\bibliographystyle{acl_natbib}

  \appendix

  \section{Results by treebank}
  \label{sec:res_by_tb}

  Results by language of the test treebanks are in Table~\ref{tab:res_all}.

\begin{table*}[th!]
\begin{tabular}{l|rrr|rrr|rrr|rrr}
        & \multicolumn{6}{c}{mBERT}                                                                                                                                                & \multicolumn{6}{c}{XLM-R}                                                                                                                                                \\
iso     & \multicolumn{1}{l}{$\phi$=0} & \multicolumn{1}{l}{$\phi$=0.5} & \multicolumn{1}{l}{$\phi$=1} & \multicolumn{1}{l}{S-P} & \multicolumn{1}{l}{S-S} & \multicolumn{1}{l}{U} & \multicolumn{1}{l}{$\phi$=0} & \multicolumn{1}{l}{$\phi$=0.5} & \multicolumn{1}{l}{$\phi$=1} & \multicolumn{1}{l}{S-P} & \multicolumn{1}{l}{S-S} & \multicolumn{1}{l}{U} \\
\toprule
aii *\#     & 8                            & \textbf{11.3}                  & 10.8                         & 1.6                     & 6.4                     & 6.0                   & 2                            & 3.3                            & 3.1                          & 2.9                     & \textbf{3.5}            & 3.1                   \\
akk *\#    & 1.5                          & 1.4                            & 1.6                          & 2.5                     & \textbf{3.0}            & 1.9                   & 2.5                          & 2.5                            & \textbf{2.8}                 & 1.9                     & 2.2                     & 2.3                   \\
am *   & \textbf{16.5}                & 10.9                           & 13.2                         & 6.6                     & 10.8                    & 10.6                  & 68.0                         & 68.6                           & 68.3                         & 68.4                    & \textbf{68.8}           & 68.1                  \\
be      & 78.5                         & 79.4                           & 79.6                         & \textbf{82.0}           & 80.9                    & 80.5                  & 85.6                         & 85.5                           & 85.6                         & 86.4                    & 86.8                    & \textbf{86.8}         \\
bho *\#    & \textbf{38.1}                & 37.8                           & 37.9                         & 37.0                    & 36.7                    & 36.7                  & 37.3                         & 37.4                           & 37.1                         & 37.4                    & \textbf{37.6}           & 37.2                  \\
bm  *\#  & \textbf{9.0}                 & 8.7                            & 8.7                          & 6.9                     & 6.7                     & 6.9                   & 6.0                          & 6.4                            & 6.2                          & \textbf{6.5}            & 6.3                     & 6.4                   \\
br      & \textbf{62.9}                & 62.6                           & 62.0                         & 60.3                    & 60.3                    & 59.6                  & 59.5                         & 59.6                           & \textbf{60.5}                & 59.9                    & 59.5                    & 58.9                  \\
bxr *\#    & 25.9                         & \textbf{26.0}                  & 25.6                         & 24.6                    & 25.5                    & 25.4                  & 27.7                         & \textbf{28.2}                  & 28.0                         & 27.2                    & 27.2                    & 26.2                  \\
cy     & \textbf{55.5}                & 55.0                           & 55.2                         & 55.1                    & 54.4                    & 54.2                  & 59.8                         & 60.1                           & 59.9                         & 60.2                    & \textbf{60.6}           & 59.6                  \\
fo  *\#    & 67.4                         & 67.8                           & \textbf{68.0}                & 66.3                    & 67.2                    & 66.4                  & 73.5                         & 72.8                           & \textbf{73.5}                & 72.6                    & 72.4                    & 73.0                  \\
gsw *\#    & 48.3                         & \textbf{48.8}                  & 48.2                         & 44.9                    & 42.2                    & 42.3                  & 46.0                         & \textbf{46.5}                  & \textbf{46.5}                & 43.6                    & 42.2                    & 44.3                  \\
gun *\#    & 8.2                          & 8.5                            & \textbf{8.7}                 & 7.3                     & 8.0                     & 8.3                   & 6.8                          & 6.8                            & \textbf{7.6}                 & 6.5                     & 5.8                     & 5.6                   \\
hsb  *\#   & 50.8                         & 51.3                           & \textbf{51.4}                & 49.4                    & 49.2                    & 49.1                  & \textbf{62.6}                & 61.9                           & 62.0                         & 61.4                    & 61.6                    & 60.0                  \\
kk      & \textbf{60.1}                & 58.9                           & 58.4                         & 58.5                    & 59.0                    & 58.2                  & 63.0                         & 62.7                           & 62.5                         & \textbf{63.7}           & 62.3                    & 61.5                  \\
kmr *    & 9.3                          & 9.2                            & 8.9                          & 8.6                     & \textbf{9.6}            & 9.5                   & \textbf{53.5}                & 53.1                           & 53.2                         & 51.8                    & 51.7                    & 52.0                  \\
koi  *\#   & 19.3                         & 18.8                           & \textbf{19.8}                & 15.8                    & 15.8                    & 16.0                  & 17.0                         & \textbf{20.1}                  & 19.1                         & 17.8                    & 17.8                    & 16.0                  \\
kpv  *\#   & 16.8                         & 17.0                           & \textbf{17.2}                & 15.6                    & 16.2                    & 15.8                  & 18.3                         & 19.1                           & \textbf{19.5}                & 17.0                    & 17.8                    & 16.3                  \\
krl  *\#   & 46.6                         & 46.4                           & 46.3                         & 46.5                    & \textbf{47.1}           & 46.4                  & 61.0                         & 61.2                           & 60.7                         & 62.0                    & \textbf{62.1}           & 61.8                  \\
mdf  *\#   & \textbf{26.1}                & 24.3                           & 24.3                         & 22.5                    & 24.5                    & 25.4                  & 20.4                         & \textbf{20.7}                  & 19.6                         & 18.4                    & 18.4                    & 16.8                  \\
mr      & 60.6                         & \textbf{61.2}                  & 60.1                         & 56.9                    & 57.7                    & 57.7                  & 69.2                         & 69.7                           & \textbf{70.0}                & 67.8                    & \textbf{70.0}           & 69.7                  \\
myv  *\#   & \textbf{20.2}                & 19.9                           & 19.8                         & 18.5                    & 19.3                    & 19.9                  & 16.8                         & \textbf{17.2}                  & 16.9                         & 16.0                    & 16.3                    & 15.5                  \\
olo *\#    & 40.7                         & \textbf{41.7}                  & 41.0                         & 41.0                    & 40.9                    & 40.5                  & 56.5                         & \textbf{56.7}                  & 56.1                         & 55.8                    & 54.3                    & 54.4                  \\
pcm  *\#   & 33.9                         & 32.8                           & 33.0                         & 32.5                    & 34.3                    & \textbf{35.4}         & \textbf{39.2}                & 39.2                           & 38.9                         & 38.0                    & 37.6                    & 37.8                  \\
sa  *    & \textbf{22.5}                & 21.9                           & 22.3                         & 21.1                    & 21.0                    & 20.6                  & 50.2                         & 49.7                           & \textbf{50.9}                & \textbf{50.9}           & 50.1                    & 50.0                  \\
ta      & 52.3                         & \textbf{54.7}                  & 54.3                         & 53.2                    & 52.0                    & 51.6                  & 54.9                         & \textbf{55.0}                  & 54.8                         & 53.8                    & 53.8                    & 54.0                  \\
te      & 69.9                         & 69.8                           & 70.0                         & 69.4                    & \textbf{70.6}           & 68.7                  & 76.0                         & 76.0                           & 76.7                         & 76.3                    & \textbf{77.1}           & 76.3                  \\
tl  \#    & 65.4                         & 57.5                           & 56.5                         & \textbf{65.8}           & 59.3                    & 65.4                  & 77.1                         & 75.7                           & 75.7                         & \textbf{78.1}           & 76.7                    & 76.4                  \\
wbp  *\#   & 5.9                          & 8.8                            & \textbf{9.2}                 & 7.5                     & 7.5                     & 7.2                   & 7.8                          & \textbf{9.5}                   & 7.5                          & 8.5                     & 5.2                     & 8.8                   \\
yo   \#   & 37.8                         & 37.9                           & 38.5                         & \textbf{39.7}           & 38.0                    & 37.5                  & 3.3                          & \textbf{3.6}                   & 3.2                          & 2.3                     & 2.7                     & 1.8                   \\
yue   *\#  & \textbf{33.0}                & 32.5                           & 32.5                         & 32.4                    & 32.4                    & 32.4                  & 41.9                         & 41.7                           & 42.0                         & \textbf{42.9}           & 42.4                    & 42.8                  \\
\bottomrule
average & \textbf{36.4}                & 36.1                           & 36.1                         & 35.0                    & 35.2                    & 35.2                  & 42.1                         & \textbf{42.3}                  & \textbf{42.3}                & 41.9                    & 41.7                    & 41.4                 
\end{tabular}

\caption{{\bf Zero-shot performance:} LAS scores on the test sets of the 30 unseen (zero-shot) languages in the language split from \citet{ustun-etal-2020-udapter} using mBERT and XLM-R. S-P=size-proportional, S-S = smooth-sampling, U=uniform. Bold indicates the best performance across models using the same PLM. * means not in mBERT and \# means not in XLM-R.}
\label{tab:res_all}
\end{table*}

  \section{Training samples}
  \label{sec:training_samples}
  The training samples are summarized in Table~\ref{tab:training_samples}.

  \begin{table*}[th!]
\begin{tabular}{l|llllllll}
  & \rotatebox[origin=c]{90}{\textsc{germanic}} & \rotatebox[origin=c]{90}{\textsc{slavic}} & \rotatebox[origin=c]{90}{\textsc{romance}} & \rotatebox[origin=c]{90}{\textsc{rom+eu}} & \rotatebox[origin=c]{90}{\textsc{rom+ar}} & \rotatebox[origin=c]{90}{\textsc{rom+tr}} & \rotatebox[origin=c]{90}{\textsc{rom+zh}} & \rotatebox[origin=c]{90}{\textsc{13lang}} \\
                             \toprule
Afrikaans-AfriBooms          & \checkmark        &       &        &           &       &       &       &       \\
Danish-DDT                   & \checkmark        &       &        &           &       &       &       &       \\
Dutch-Alpino                 & \checkmark        &       &        &           &       &       &       &       \\
English-EWT                  & \checkmark        &       &        &           &       &       &       & \checkmark      \\
German-HDT                   & \checkmark        &       &        &           &       &       &       &       \\
Gothic-PROIEL                & \checkmark        &       &        &           &       &       &       &       \\
Icelandic-IcePaHC            & \checkmark        &       &        &           &       &       &       &       \\
Norwegian-Bokmaal            & \checkmark        &       &        &           &       &       &       &       \\
Swedish-Talbanken            & \checkmark        &       &        &           &       &       &       & \checkmark      \\
Czech-PDT                    &         & \checkmark      &        &           &       &       &       &       \\
Old\_Church\_Slavonic-PROIEL &         & \checkmark      &        &           &       &       &       &       \\
Old\_Russian-TOROT           &         & \checkmark      &        &           &       &       &       &       \\
Polish-LFG                   &         & \checkmark      &        &           &       &       &       &       \\
Russian-SynTagRus            &         & \checkmark      &        &           &       &       &       & \checkmark      \\
Serbian-SET                  &         & \checkmark      &        &           &       &       &       &       \\
Slovak-SNK                   &         & \checkmark      &        &           &       &       &       &       \\
Ukrainian-IU                 &         & \checkmark      &        &           &       &       &       &       \\
French-GSD                   &         &       & \checkmark       & \checkmark          & \checkmark      & \checkmark      & \checkmark      &       \\
Italian-ISDT                 &         &       & \checkmark       & \checkmark          & \checkmark      & \checkmark      & \checkmark      & \checkmark      \\
Portuguese-GSD               &         &       & \checkmark       & \checkmark          & \checkmark      & \checkmark      & \checkmark      &       \\
Romanian-RRT                 &         &       & \checkmark       & \checkmark          & \checkmark      & \checkmark      & \checkmark      &       \\
Spanish-AnCora               &         &       & \checkmark       & \checkmark          & \checkmark      & \checkmark      & \checkmark      &       \\
Basque-BDT                   &         &       &        & \checkmark          &       &       &       & \checkmark      \\
Arabic-PADT                  &         &       &        &           & \checkmark      &       &       & \checkmark      \\
Chinese-GSD                  &         &       &        &           &       &       & \checkmark      & \checkmark      \\
Turkish-IMST                 &         &       &        &           &       & \checkmark      &       & \checkmark      \\
Finnish-TDT                  &         &       &        &           &       &       &       & \checkmark      \\
Hebrew-HTB                   &         &       &        &           &       &       &       & \checkmark      \\
Hindi-HDTB                   &         &       &        &           &       &       &       & \checkmark      \\
Japanese-GSD                 &         &       &        &           &       &       &       & \checkmark      \\
Korean-GSD                   &         &       &        &           &       &       &       & \checkmark     
\end{tabular}
\caption{Treebanks included in the different samples}
\label{tab:training_samples}
\end{table*}

\section{Results by treebank with the different samples}
\label{sec:tb_res_sample}

\begin{figure*}[t]
  \hspace{-0.4cm}
  \includegraphics[scale=0.85]{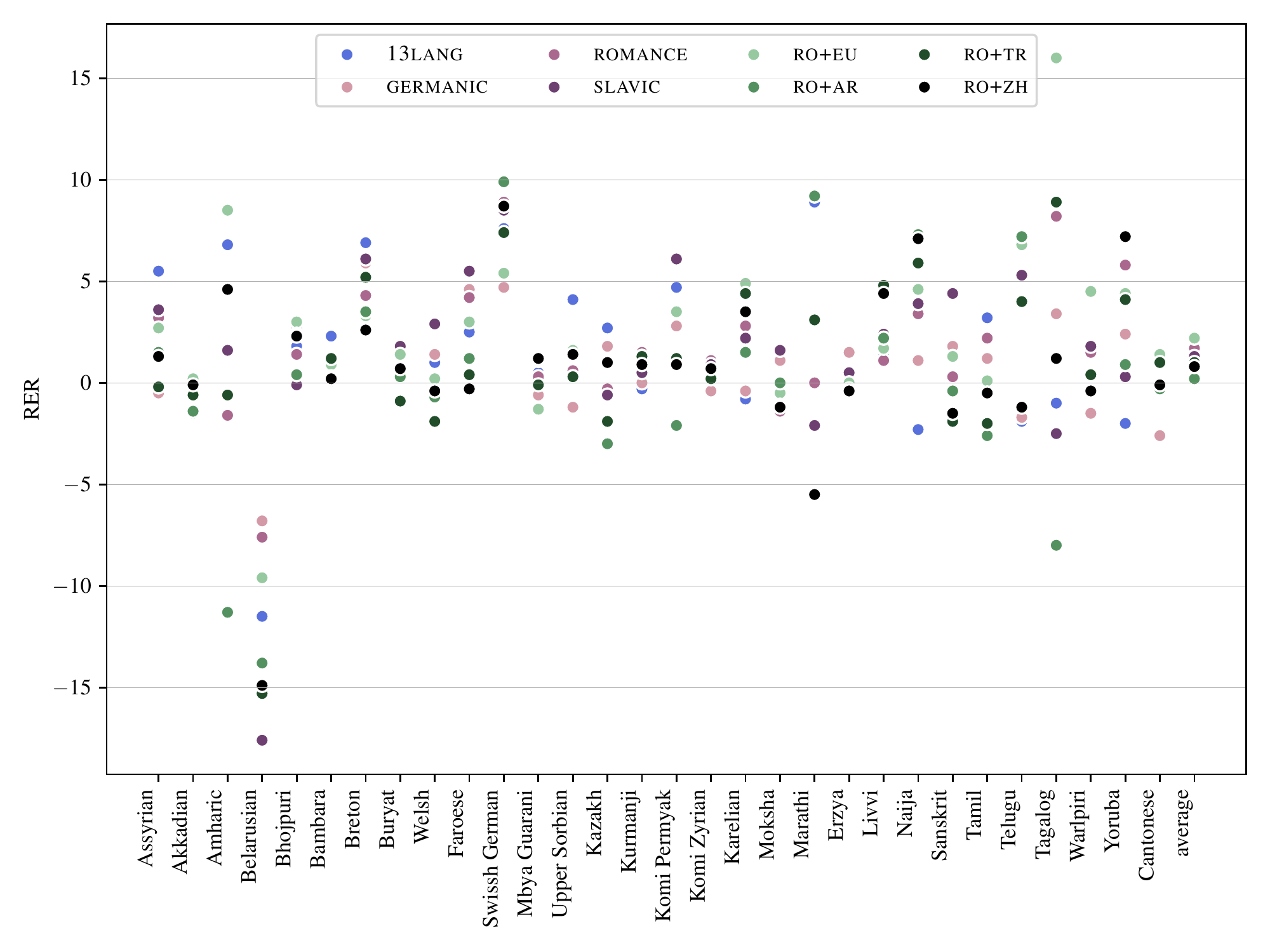}
    \caption{Relative error reduction (RER) in LAS points between our best worst-case aware result and the best baseline for each training sample used on test sets in the 30 languages.}
    \label{fig:tb_res_plot}
\end{figure*}

  Relative error reduction between our best worst-case aware result and the best baseline for each training sample used, with mBERT, in Figure~\ref{fig:tb_res_plot}.

\end{document}